\documentclass[letterpaper, 10 pt, conference]{ieeeconf}  

\usepackage{etoolbox}
\makeatletter
\patchcmd{\@makecaption}
  {\scshape}
  {}
  {}
  {}
\makeatother

\IEEEoverridecommandlockouts                              

\usepackage{graphicx}
\usepackage{graphics} 
\usepackage{epsfig} 
\usepackage[backend=bibtex]{biblatex}
\usepackage{amsmath} 
\usepackage{hyperref}
\usepackage{adjustbox}
\usepackage{multirow}
\usepackage{boldline}
\usepackage{comment}
\usepackage{times} 
\usepackage{amssymb}  
\usepackage{algorithmic}
\usepackage{algorithm}
\usepackage{pseudo}
\usepackage{xcolor}
\usepackage{array}
\newcolumntype{P}[1]{>{\centering\arraybackslash}p{#1}}

\hypersetup{
     colorlinks   = true,
     citecolor    = orange
}

\title{\LARGE \bf
Bird's Eye View Based Pretrained World model for Visual Navigation
}

\author{Anonymous Authors}
\author{Kiran Lekkala$^{\textbf{*}}$, Chen Liu$^{\textbf{*}}$ and Laurent Itti$^{\dagger}$
\thanks{$^{*}$\textbf{Equal Contribution}}
\thanks{$^{1}$ The authors are with Thomas Lord Department of Computer Science, University of Southern California, 90089, USA
        {Correspondence to \tt\small klekkala@usc.edu}}%
\thanks{This work was supported by the National Science Foundation (award 2318101), C-BRIC (one of six centers in JUMP, a Semiconductor Research Corporation (SRC) program sponsored by DARPA) and the Army Research Office (W911NF2020053). The authors affirm that the views expressed herein are solely their own, and do not represent the views of the United States government or any agency thereof.}
}

\bibliography{paper}
\begin{document}

\maketitle
\thispagestyle{empty}
\pagestyle{empty}

\begin{abstract}

Sim2Real transfer has gained popularity because it helps transfer from inexpensive simulators to real world. This paper presents a novel system that fuses components in a traditional World Model into a robust system, trained entirely within a simulator, that Zero-Shot transfers to the real world. To facilitate transfer, we use an intermediary representation that is based on \textit{Bird's Eye View (BEV)} images. Thus, our robot learns to navigate in a simulator by first learning to translate from complex \textit{First-Person View (FPV)} based RGB images to BEV representations, then learning to navigate using those representations. Later, when tested in the real world, the robot uses the perception model that translates FPV-based RGB images to embeddings that were learned by the FPV to BEV translator and that can be used by the downstream policy.
The incorporation of state-checking modules using \textit{Anchor images} and Mixture Density LSTM not only interpolates uncertain and missing observations but also enhances the robustness of the model in the real-world. We trained the model using data from a Differential drive robot in the CARLA simulator. Our methodology's effectiveness is shown through the deployment of trained models onto a real-world Differential drive robot. Lastly we release a comprehensive codebase, dataset and models for training and deployment (\url{https://sites.google.com/view/value-explicit-pretraining}).
\end{abstract}

\section{Introduction}


\textit{Reinforcement Learning (RL)} has predominantly been conducted in simulator environments, primarily due to the prohibitive costs associated with conducting trial-and-error processes in the real world. With the advances in graphics and computational technologies, there has been a significant development in realistic simulators that capture the system (robot) information. However, the domain gap between synthetic and real data introduces a substantial performance drop when the models are directly deployed into real-world applications after training, a phenomenon commonly referred to as the \textit{Sim2Real gap}.

 Traditionally, Sim2real transfer methods either optimize training on a simulation that closely resembles real-world data, or use \textit{Domain Randomization} \cite{tobin2017domain} or \textit{Domain Adaptation} \cite{Truong_2021}. Other works \cite{DBLP:conf/icra/BewleyRLHSLK19} train on a simulated environment and deploy to self-driving cars. However, since these models were not trained in cluttered, pedestrian-rich environments, they would not generalize to some real-world scenarios. Some of the recent works, such as \cite{DBLP:conf/icra/SteinR18} and \cite{DBLP:journals/ral/ZhangTYX0BB19}, have shown promising results in attempting to cope with the Sim2real gap using \textit{Style Transfer} for data generation using limited real-world data but do not focus on learning optimal representations for performing the navigation task. On the other hand, models that are trained end-to-end in a simulator overfit to the trained task, without learning generalizable representations \cite{DBLP:journals/ral/AminiGPMBKR20}. With all these practical considerations, it is imperative that we design a robust and low resource-footprint model that enables a mobile-robot to efficiently function in diverse scenarios.



\begin{figure}
\centering
    \includegraphics[width=0.85\columnwidth]{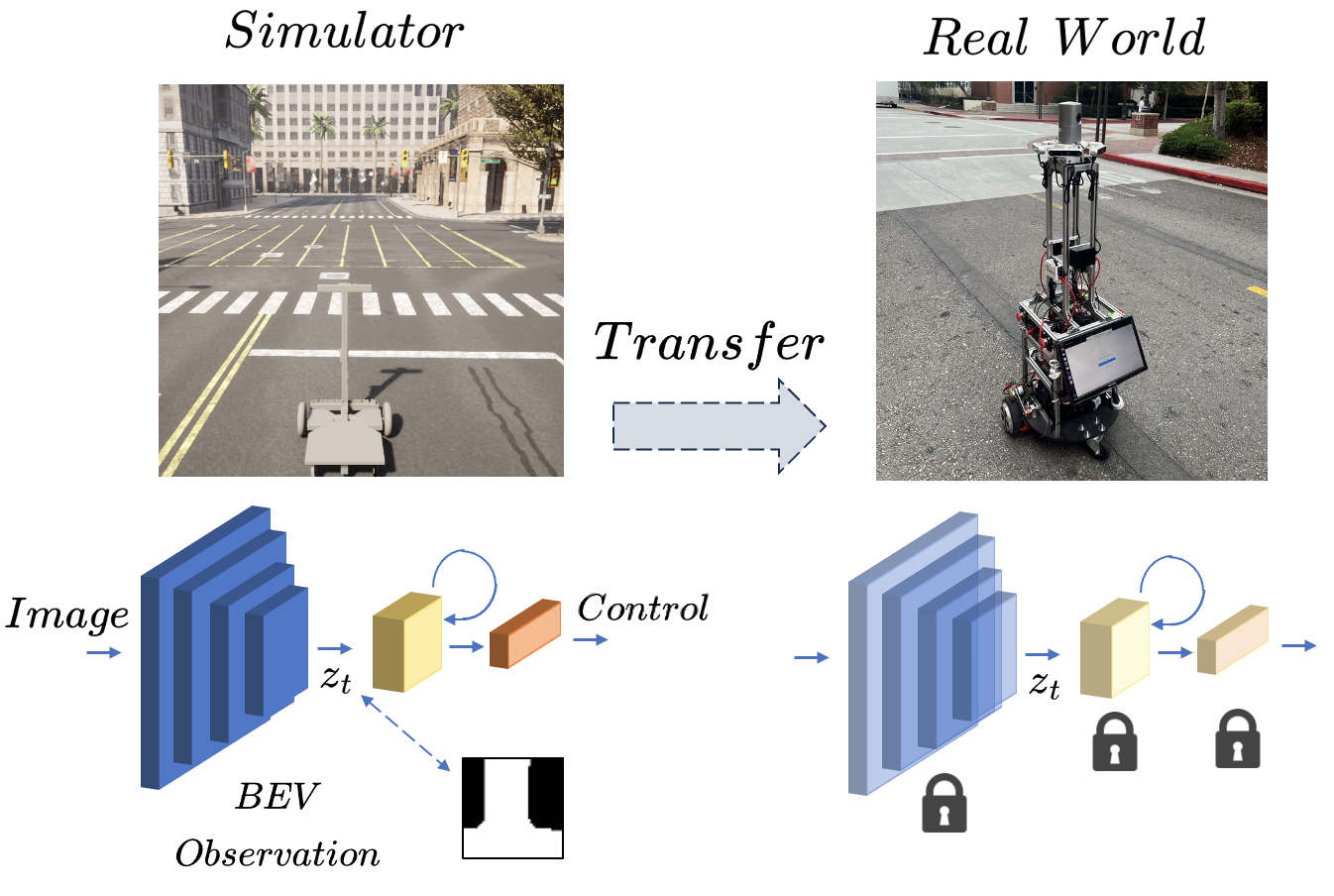}
    \caption{\textbf{Overview of our system} We first train the visual navigation system on a large-scale dataset collected in the simulator and deploy the frozen model in an unseen real-world environment.}
    \label{fig:introsim}
\end{figure}

In this paper, we formulate a new setting for \textit{Zero-shot} Sim2Real transfer for Visual Navigation without Maps, involving data obtained from the CARLA simulator, as outlined in Fig. \ref{fig:introsim}. To avoid any Sim2Real gap within the control pipeline and focus only on the perception transfer, we built a Differential-drive based robot in the CARLA simulator that closely resembles our real-world robot. Using this setup, we build a large dataset consisting of \textit{First-person view (FPV)} and \textit{Bird's eye view (BEV)} image sequences from the CARLA \cite{carla} simulator. The system is trained entirely on this simulated dataset and is frozen and deployed on a real-world mobile robot. 

\begin{figure}
    \centering
    \includegraphics[width=\columnwidth]{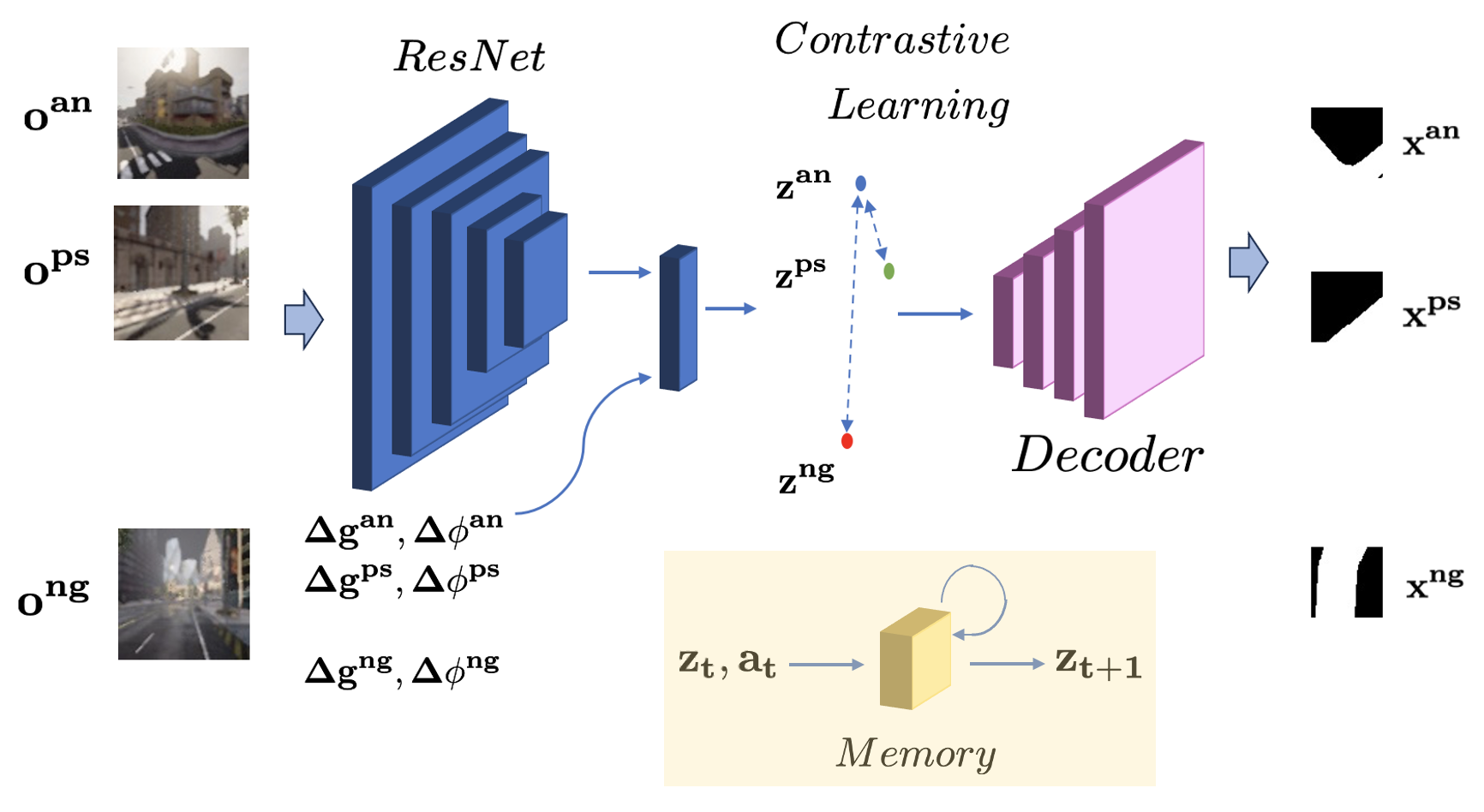}
    \caption{\textbf{Training pipeline for the perception model}. (a) During the training phase, the ResNet model is trained using a set of temporal sequences, consisting of pairs of input (FPV images, displacement and orientation to goal) and output (BEV images) from the simulator. Our contrastive loss embeds positive closer to anchor and negative farther away. (b) In the bottom, we pictorially show the input and the output that is used to train the memory module.}
    \label{fig:3}
\end{figure}


\section{Related work}

Although, many methods \cite{DBLP:journals/corr/abs-2305-15591, kaspar2020sim2real, DBLP:conf/icra/LekkalaI21, lekkala2015artificial} use simulators for learning through an extensive amount of experiences that could be used to train a model policy end to end, some recent works \cite{arndt2019meta} have shown promising results, on various tasks \cite{josifovski2022analysis,}, using encoders that are pretrained on large unlabelled expert data and then train a significantly smaller network on top of the frozen encoder. Since these encoders are not trained on a specific task, we call it pretraining. Representations estimated using these \textit{pretrained and frozen} encoders would help the model remain lightweight and flexible, which is desirable for mobile platforms. In our work, we employ such an approach with a new pre-training objective (to reconstruct BEV maps from FPV inputs), which we show provides very good generalization for downstream robotics tasks. Since learning representations does not involve any dynamics, any navigation dataset consisting of FPV-BEV could be used to pretrain the encoder. By training a Vision encoder using a large aggregated dataset, this could be a comparable alternative to the current ViT's \cite{DBLP:conf/icml/RadfordKHRGASAM21} used for Robotics.





 \textit{Bird's Eye View (BEV)} based representation allows for a compact representation of the scene, invariant to any texture changes, scene variations, occlusions or lightning differences in an RGB image. This makes for an optimal representation for \textit{PointGoal Navigation}. Furthermore, it is one of the most efficient and lightweight form of information, since the BEV maps are binary. For example, the corresponding BEV image of an 1MB FPV image is around 0.5KB. Some works estimate BEV maps from RGB images, such as \cite{Lu_2019}, \cite{Pan_2020} and \cite{reiher2020sim2real}. However, these map predictions from FPV images are typically only evaluated for visual tasks, with a lack of evidence that BEV-based representations can be useful for robotic tasks. Furthermore, \cite{DBLP:conf/nips/AnandROBCH19} have shown that reconstruction-based methods like VAE \cite{DBLP:journals/corr/KingmaW13} perform close to Random encoders. Incorporating these representations as inputs for training downstream models for robotic tasks to ensure their compatibility indeed is challenging. Our pretraining approach not only allows for learning visual representations that are optimal for robotic tasks, but also allows these representations to reconstruct the corresponding BEV map. Together, they allow the lightweight policy model to efficiently learn the task through these representations.

 




\textit{Recurrent world-models.} \cite{DBLP:conf/nips/HaS18} introduces a novel approach to RL, incorporating a vision model for sensory data representation and a memory model for capturing temporal dynamics, all of which collectively improve agent performance. Apart from the advantages of pertaining each module, some of the modules in this architecture can be frozen after learning the representation of the environment, paving the way for more efficient and capable RL agents.



We propose a novel training regime and develop a perception model pretrained on a large simulated dataset to translate FPV-based RGB images into embeddings that align with the representations of the corresponding BEV images. Along with that, we upgrade the existing world models framework using a novel model-based \textit{Temporal State Checking (TSC)} and \textit{Anchor State Checking (ASC)} methods that add robustness to the navigation pipeline when transferred to the real world. We release the code for pre-training, RL training and ROS-based deployment of our system on a real-world robot, FPV-BEV dataset and pre-trained models. With the above contributions, we hope move closer towards open-sourcing a robust Visual Navigation system that uses pre-trained models trained on large datasets and simulations for efficient representation learning.

\begin{figure*}
\centering
    \includegraphics[width=.75\textwidth]{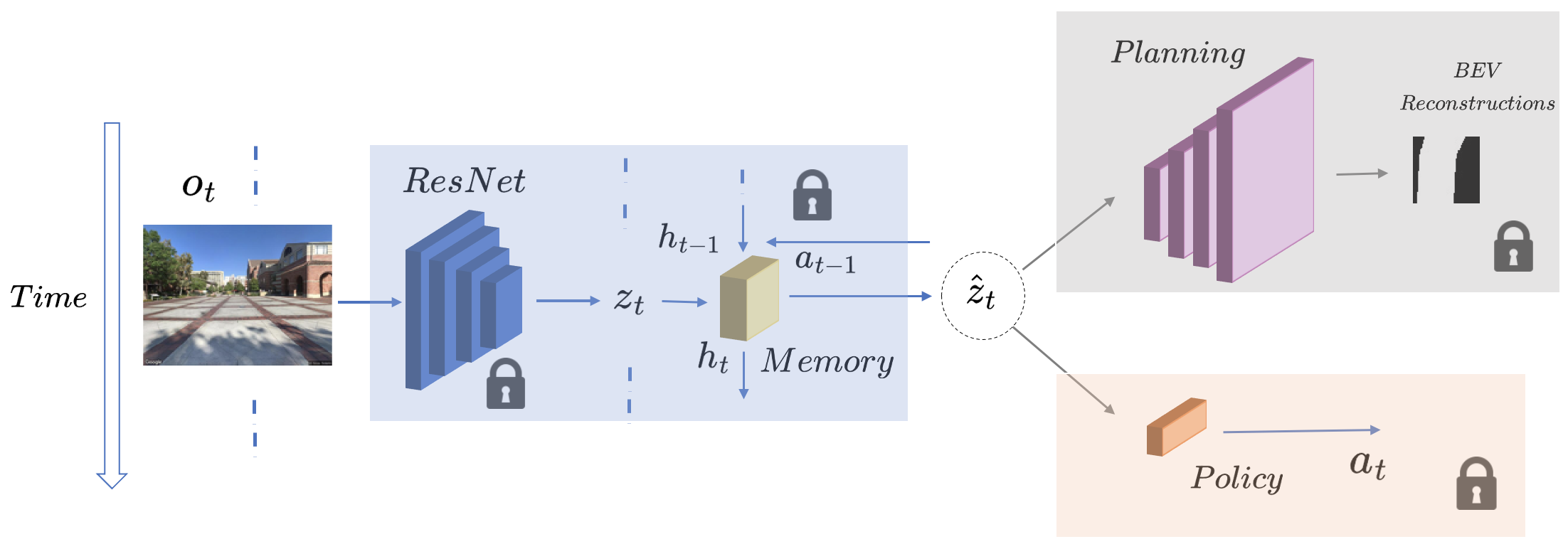}
    \caption{\textbf{Working of the System.} RGB observation $o_t$ at time step $t$ is passed to the perception model (\textcolor{blue}{blue}) that compresses it into an embedding $z_t$. The memory model takes the current latent representation $z_t$ and uses the historical context to refine the state into $\hat{z}_t$. These embeddings could either be used to train a control policy (\textcolor{orange}{orange}) or to reconstruct the  Bird's Eye View (BEV) for planning (\textcolor{gray}{grey}). Both utilities result in an action command $a_t$.}
    \label{fig:dyn}
\end{figure*}

\section{Proposed Method}

For an autonomous agent to navigate using camera imagery, we use a simple system that consists of a perception model and a control model as shown in \ref{fig:dyn}. The perception model takes input observation $o_t$ and outputs an embedding $z_t$ that is then passed on to the policy, as part of the control model to output an action vector $a_t$, throttle and steer. We first outline the perception model, with the objective of efficiently learning compact intermediate representations compatible with downstream policy learning, solely from a sequence of observations from the simulator. We then describe our second contribution, which involves the enhancement of the robustness and stability of the predictions during real-world evaluation.

\subsection{Perception model}

When training the perception model, we focus on 3 main principles. Firstly, $o_t$, the embedding vector should always be consistent with the BEV reconstruction. Secondly, BEV images must be represented in a continuous latent space that has smooth temporal transitions to similar BEV images. Finally, the perception model must efficiently utilize an unlabelled sequence of images as an expert video portraying optimal behaviour. This would also allow for unsupervised training/fine-tuning of the model using real-world expert videos, which we leave for future work.

The perception model consists of a \textit{ResNet-50} \cite{resnet} that is tasked with processing the observation $o_t$ obtained from an RGB camera, with the primary objective of comprehending the environmental context in which the robot operates, and compresses $o_t$ into a consistent intermediate representation, $z_t$, which when decoded through a BEV decoder, outputs a BEV image $x_t$. Our choice for BEV observations is rooted in their capacity to convey the surrounding roadmaps with minimal information redundancy. To learn such representations from a set of FPV and corresponding binary BEV images, prior methods \cite{Lu_2019} train a \textit{Variational Autoencoder (VAE)} \cite{DBLP:journals/corr/KingmaW13} to encode an RGB image $o_t$ that is decoded using $z_t \in \mathbb{R}^{B \times d }$, where $B$ is the batch size and $d$ is the embedding dimension. Given that we have batches $\mathbf{x}$ (BEV reconstructions) and $\mathbf{y}$ (ground-truth BEV observations), we could then optimize the following reconstruction loss $\mathcal{L}_R$:


\vspace{-5pt}
\begin{equation}
\mathcal{L}_{R} = -[\mathbf{y} \cdot log(\mathbf{x}) + (1 - \mathbf{y}) \cdot log(1 - \mathbf{x})]
\end{equation}

Using the above loss, the VAE Encoder will learn to embed the FPV observations $\mathbf{o}$ that will reconstruct their corresponding BEV observations $\mathbf{x}$, and $\mathbf{y}$ being the corresponding BEV labels. Additionally, $KL$ (\textit{Kullback Leibler}) divergence forces the embeddings, to be within a Gaussian distribution of zero-mean and unit-covariance, that allows for smooth interpolation. The representations learnt by VAE would embed 2 FPV observations that are very similar, for example, 2 straight roads, but a have slight variation in the angle to be closer, than a straight road and an intersection. The following is the loss function used to train a VAE.

\begin{equation}
\label{eqn:elbo}
\mathcal{L}_{ELBO} =  \mathcal{L}_{R} + \beta \cdot KL[\mathcal{N}(\mathbf{\mu}, \mathbf{\sigma}^2) \mid\mid \mathcal{N}(0, 1)]
\end{equation}

Although, the above ELBO loss would allow the model to learn appropriate representations for understanding the observation, these representations do not capture the temporal understanding of the task. Typically, representations for robotics embed observations in such a way to make it easier for the policy to learn the behaviour of an objective quickly and efficiently. One of the earliest methods for self-supervised learning, Time-Contrastive Networks \cite{DBLP:conf/icra/SermanetLCHJSLB18} disambiguates temporal changes by embedding representations closer in time, closer in the embedding space and farther otherwise by optimizing the following loss function.

\begin{align}
\label{eqn:tcn}
\begin{split}
\mathcal{L}_{InfoNCE} = \mathbb{E}_{\mathbf{z}^{ps}}\left[-\log \frac{\mathcal{S}_\phi \left(\mathbf{z}^{an}, \mathbf{z}^{ps}\right)}{\mathbb{E}_{\mathbf{z}^{ng}} \mathcal{S}_\phi\left(\mathbf{z}^{an}, \mathbf{z}^{ng}\right)}\right]
\end{split}
\end{align}

In the above function, $\mathbf{z^{an}}$, $\mathbf{z^{ps}}$ and $\mathbf{z^{ng}}$ are a batch of embeddings corresponding to anchors, positives and negatives and $\mathcal{S_\phi}$ is the similarity metric of the embeddings from the encoder $f_\phi$. For a given single observation sample $o_t$, the embedding obtained as an anchor $z^{an}$, we uniformly sample a frame within a temporal distance threshold $d_{thresh}$ to obtain $z^{ps}$ at timestep $t+\delta$ and $z^{ng}$, anywhere from $t+\delta$ to the end of the episode. However, recently \cite{DBLP:conf/iclr/MaSJBK023} has shown that in-domain embeddings learnt by TCN are discontinuous, leading to sub-optimal policies. To alleviate this problem, we also add the reconstruction loss $\mathcal{L}_R$ that enhances the stability of the training process, and helps learn better representations. To achieve the FPV-BEV translation using our method, we optimize the model parameters using the following contrastive with reconstruction loss $\mathcal{L}_{CR}$ for image encoding.

\begin{equation}
\label{eqn:cr}
\mathcal{L}_{CR} = \mathcal{L}_{R} + \beta \cdot \mathcal{L}_{InfoNCE} 
\end{equation}

In the above loss function, $\beta$ balances the reconstruction with the contrastive loss, since the model optimizes the reconstruction loss slower than the contrastive loss. Using the above loss function, the model learns more temporally continuous and smoother embeddings as it constrains the proximity of the embeddings not only using the contrastive learning loss but also based on the BEV reconstructions.

\subsection{Temporal model with Robustness modules}

\begin{figure}
\centering
    \includegraphics[width=.85\columnwidth]{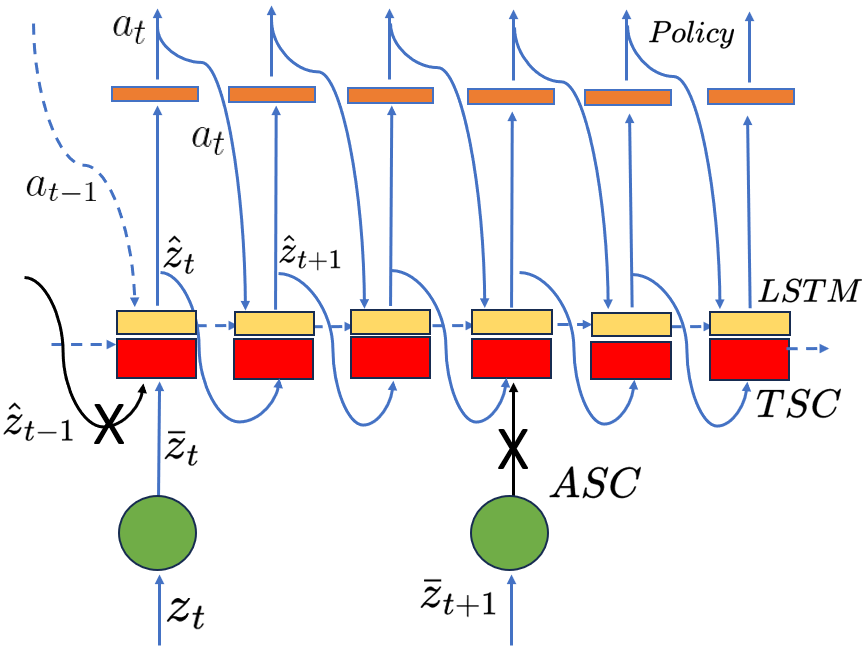}
    \caption{\textbf{Robustness enhancement using Memory module.} \textit{TSC} (\textcolor{red}{red}) only takes input from the representation $z_t$ when it comes with a high confidence score. Otherwise, it takes the previous prediction by the LSTM $\hat{z}_{t-1}$ as interpolation. \textit{ASC} (\textcolor{green}{green}) improves the representation of the incoming observation by making it in-domain. The crosses above correspond to rejecting the precepts and using the model's state prediction as the current state.}
    \label{fig:5}
\end{figure}

To enhance the robustness of the perception model and transfer it to the real world setting, we implemented an additional model in the pipeline. Fig. \ref{fig:5} shows our proposed method of robustness enhancement. This involves the integration of an LSTM, functioning as a \textit{Memory} model. The LSTM was trained on sequences $\{\langle o_j, a_j \rangle\}_{j=0}^{j=T}$ gathered from sequences $\{\mathcal{T}_0, \mathcal{T}_1, .. , \mathcal{T}_n\}$ in the simulator. The primary outcome of this Memory model is to effectively infuse historical context $\{\langle z_j, a_j\rangle\}_{j=0}^{j=T}$ into the prediction of $\hat{z}_t$, which forms a candidate of $z_t$, and enhancing the robustness of the perception module when confronted with the unseen real-world data. To model the uncertainty in future states, we add an \textit{Mixture Density Network (MDN)} on the top of LSTM output layer. The above pipeline is formulated as:

\begin{equation}
    \hat{z}_t \sim P(\hat{z}_t \mid a_{t-1}, \hat{z}_{t-1}, h_{t-1})
\end{equation}

where $a_{t-1}, \hat{z}_{t-1}, h_{t-1}$ respectively denotes action, state prediction at the previous timestep, and historical hidden state at the time step $t-1$. $\hat{z}_t$ is the latent representation that is given as an input to the policy. We optimize M with the below loss function:
\begin{equation}
    \mathcal{L}_{M} = -\frac{1}{T}\sum^T_{t=1}{\rm log}(\sum^K_{j=1} \mathbf{\theta} \cdot \mathcal{N}(z_t | \mu_j, \sigma_j))
\end{equation}

where $\{T, K, \theta_j, \mathcal{N}(z_t | \mu_j, \sigma_j) \}$ is, respectively, the training batch size, number of Gaussian models, Gaussian mixture weights with the constraint $\sum^K_{j=1}\theta_j=1$, and the probability of ground truth at time step $t$ conditioned on predicted mean $\mu_j$ and standard variance $\sigma_j$ for Gaussian model $j$.

Nonetheless, it is noteworthy that $z_t$ that is obtained from the ResNet-50 may be slightly distinct from the latent distribution of BEV images when the perception model is applied to real-world observations $o_t$, potentially impacting the performance of the LSTM and the policy. To mitigate this concern, we collected a dataset $\mathcal{R}$ comprising of the BEV-based latent embeddings $s \in \mathcal{R}$ of 1439 FPV images which we define as the \textit{BEV anchors}. In practice, upon obtaining the output vector $z_t$ from the ResNet-50, we measure its proximity to each $s \in \mathcal{R}$, subsequently identifying the closest match. We replace $z_t$ with the identified anchor embedding $\bar{z}_t$, ensuring that both the LSTM and the policy consistently uses the pre-defined BEV data distribution. We pass $\bar{z}_t$ as an input to the LSTM, along with the previous action $a_{t-1}$ to get the output $\hat{z}_{t+1}$. Again, we find the closest match $\hat{s}_t \in \mathcal{R}$ for $\hat{z}_t$. We call this module \textit{Anchor State Checking (ASC)}:
\begin{equation}
\bar{z} = \arg\min_{s \in S} \| z - s \|
\end{equation}

\begin{figure}
\centering
    \includegraphics[width=.9\columnwidth]{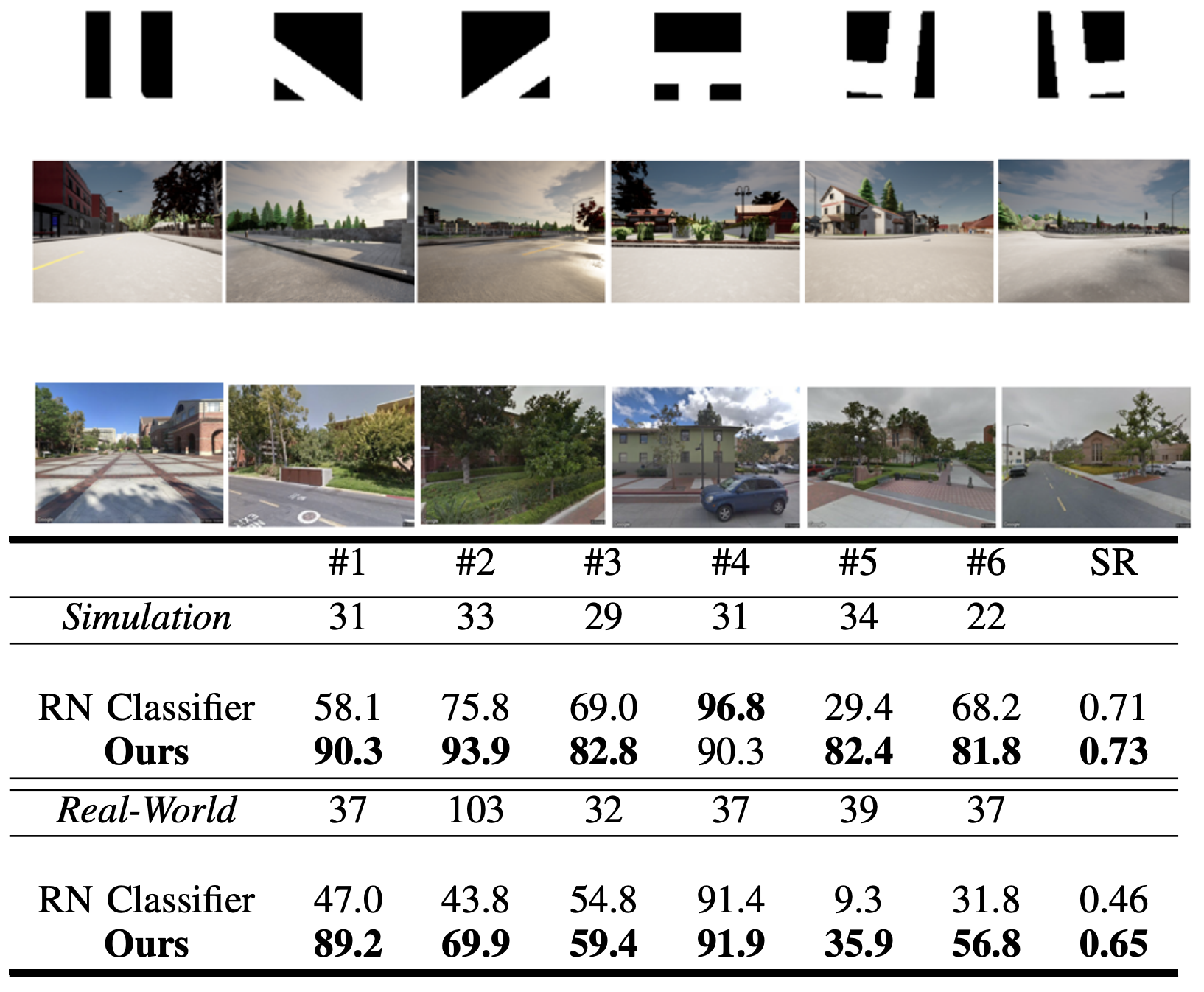}
    \caption{\textbf{Out-of-domain and real-world evaluation} We constructed two 6-class validation datasets: one from the simulator (first row in the table) and another from street-view data (second row). Each class corresponds to the BEV images shown above. We specify accuracies for each class. Along with that, we also specify the success rate (SR) of the agent, when the encoder is deployed for real-world visual navigation. Our method outperformed the ResNet classifier (baseline) on both the unseen simulation dataset, the real-world validation dataset and real-world navigation as shown above.}
    \label{fig:table_qual}
\end{figure}
We also utilize the LSTM model for rejecting erroneous predictions by the ResNet-50, further enhancing the system's robustness against noise. If the  processed prediction $\Bar{z}_t$ from the perception model is estimated with confidence score $\tau_t$, obtained from either cosine-similarity or MSE, below a predefined threshold $\rho$, we deliberately discard $\Bar{z}_t$ and opt for $\hat{z}_t$. In such instances, we resort to the output of the LSTM at the previous time-step. This module is known as \textit{Temporal State Checking (TSC)}:

\begin{equation}
\label{eq6}
\hat{z}_t =\left\{
\begin{aligned}
\Bar{z}_t & , & \tau_t \geq \rho, \\
\hat{z}_{t-1} & , & \tau_t < \rho.
\end{aligned}
\right.
\end{equation}

Apart from adding robustness to the system using TSC, the utilization of the Memory model also serves as the crucial purpose of performing interpolation for the robots state in instances where actual observations $o_t$ are delayed, ensuring the continuity and reliability of the entire system. There is often a notable discrepancy in the update frequencies between control signals and camera frames, since control signals often exhibit a significantly higher update rate (50Hz) compared to the incoming stream of camera frames (15Hz). Values mentioned in brackets is in regards to our setup. This is also beneficial in the case of the recent large vision-language models like RT-X \cite{DBLP:journals/corr/abs-2310-08864} that could solve many robotic tasks, but with a caveat of operating at a lower frequency, typically around 5Hz.

\begin{figure}
\centering
    \includegraphics[width=\columnwidth]{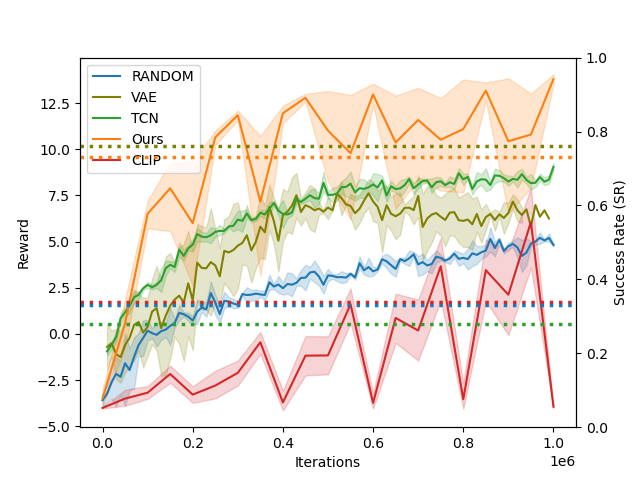}
    \caption{\textbf{Policy learning and Planning experiments on navigation task using pretrained representations}. Using a pretrained \textit{ResNet} encoder, we compare our method with different baselines. The training curves are obtained when we train a 1-layer policy, using RL, that takes the embeddings from the frozen encoder. The $x$ and $y$ axis corresponds to iterations and the cumulative reward, with the shaed regions showing the 95\% confidence intervals. We also perform planning experiments, where the BEV reconstructions are used to navigate to the goal, as shown the by the success rate (SR), through the dotted lines corresponding to each method.}
    \label{fig:6}
\end{figure}

\section{Experimental platform and setup}

To leverage the extensive prior knowledge embedded in a pre-trained model, we opt to train a ResNet-50 \cite{resnet} model after initializing with ImageNet pre-trained weights on a large-scale dataset containing FPV-BEV image pairs captured in the simulator. We collected the train dataset from the CARLA simulator to train both the Perception and the Memory model. Along with that, we also collected the validation and the test datasets from 2 different real-world sources. Following are the details on the collected datasets.

\begin{figure*}
\centering
    \includegraphics[width=.85\textwidth]{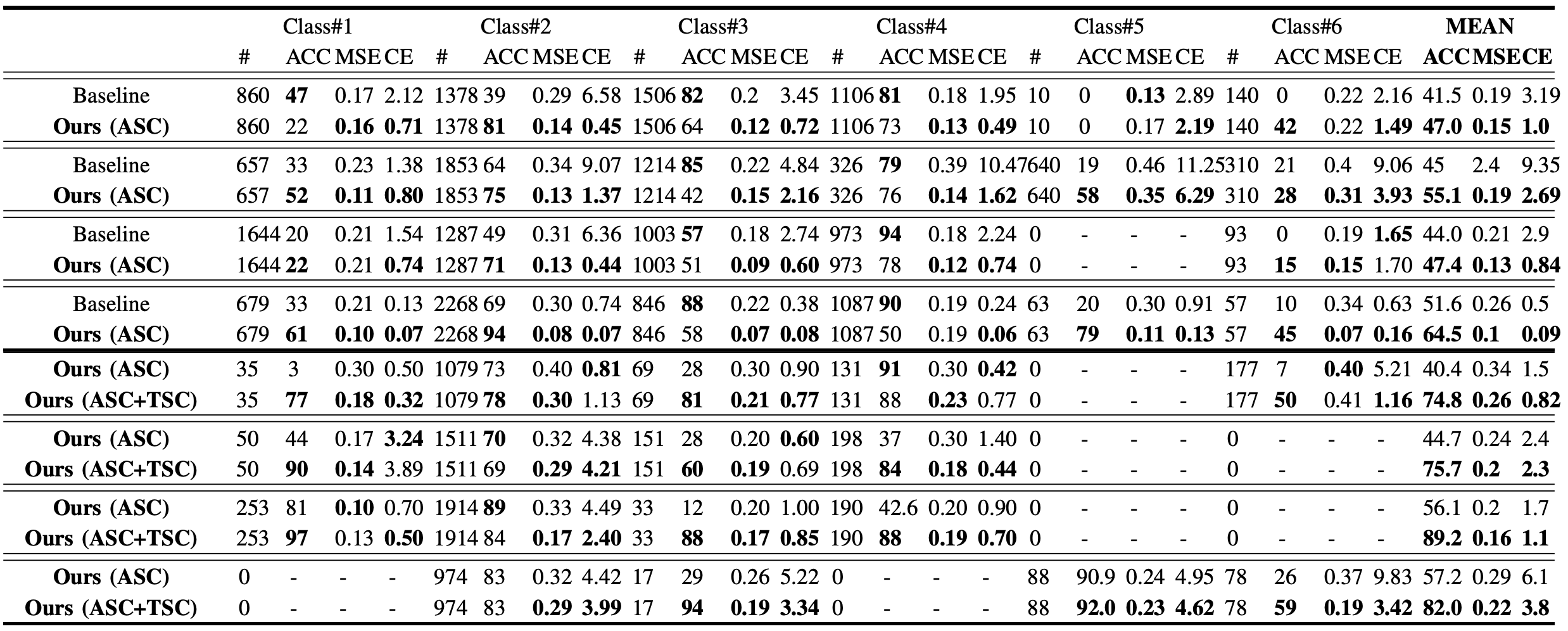}
    \caption{\textbf{Ablation experiments on the Test Dataset.} Each double-row corresponds to a data sequence. We demonstrate that our approach not only attains high ACC (accuracy), but also provides a more granular BEV representation compared to the naive classifier, as indicated by the MSE (Mean Squred Error) and CE (Cross-Entropy) metrics. In the upper portion of the table, we assessed our method independently of the LSTM on an unseen temporal sequence from the simulator, contrasting it with the baseline CNN classifier. In the lower portion, we compared the performance of system with and without LSTM on a real-world data sequence. Note that dashes in the table indicate the absence of a class in the respective sequence. We compute the mean values for each row as shown in the last column.}
    \label{fig:table_ablation}
\end{figure*}
\subsection{Experimental platform}

For evaluating Zero-shot Sim2Real transfer, we built a hardware apparatus which is a Non-Holonomic, Differential-drive robot (\textit{Beobotv3}) for the task of visual navigation. Our system is implemented using the \textit{ROS} (Robotic Operating System) middleware and uses a \textit{Coral EdgeTPU}, which is an ASIC chip designed to run CNN models for edge computing for all the compute. We used this Edge-TPU to run the forward inference of the ResNet-50 through a ROS nodes.


The CARLA simulator had been primarily tailored to self-driving applications, that use \textit{Ackermann steering}, we further developed an existing differential drive setup using \textit{Schoomatic} \cite{Scoomatic} and upgraded the CARLA simulator. We find this necessary because our real-world hardware system is based on differential-drive and to enable seamless transfer without any Sim2Real gap in the control pipeline, both the control systems need to have similar dynamics. In response to this limitation, Luttkus \cite{Scoomatic} designed a model for the integration of a differential-drive robot into the CARLA environment. Building upon their work, we undertook the development of a version of CARLA simulator catering to differential-drive robots for reinforcement learning, subsequently migrating it into the newly introduced CARLA \textit{0.9.13}.






\subsection{Data collection}


\subsubsection{Train dataset from CARLA simulator}
Within the CARLA simulator, we have access to the global waypoints along various trajectories. To allow more diversity, we randomly sampled a range of different orientations and locations. Leveraging this setup, we facilitated the generation of a large dataset of FPV-BEV images. We augmented the simulator's realism by introducing weather randomization and non-player traffic into the simulated environment.

\subsubsection{Validation dataset from Google Street View}
Using the Google Street View API, we obtained all the panoramic images from various locations on the USC campus. The panoramic images were segmented with a Horizontal Field of View (FoV) of 90 degrees and are manually segregated into different 6 different classes as shown in Fig. \ref{fig:table_qual}. The validation dataset does not have any temporal sequencing and is primarily focused on having a broader and more uniform data distribution across all the classes. Due to these reasons, this dataset becomes an optimal choice for evaluating the perception model.


\subsubsection{Test dataset from Beobotv3}
To evaluate the quality of representations estimated by the entire system, we record a video sequence using a mobile robot. More precisely, we recorded a set of 5 \textit{ROSBag} sequences at different locations of the USC campus. Later, we labelled all the frames in a \textit{ROSBag} sequence, similar to the above paragraph. However, unlike the validation set, the test dataset has temporal continuity, which helps us judge the entire navigation system.

\begin{figure}
\centering
    \includegraphics[width=.8\columnwidth]{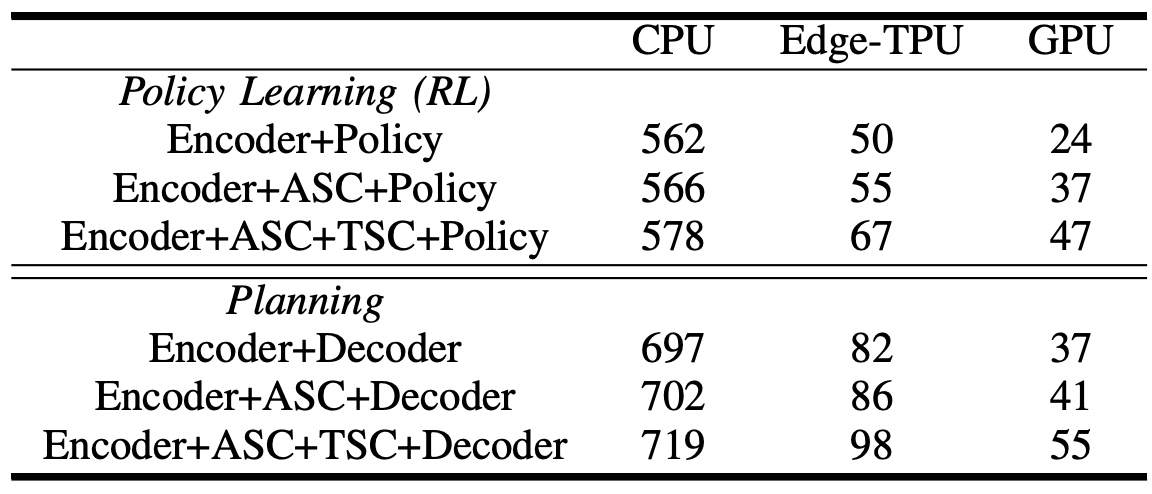}
    \caption{\textbf{Comparison of runtime}. Computation costs (runtime in \textit{milliseconds}) of each module in the navigation system for policy learning and planning are shown above.}
    \label{fig:table_flops}
\end{figure}

\section{Evaluation and Results}

Through our experiments, we aim to answer the following questions in regards to our proposed method.

\begin{enumerate}
    \item \textit{How good are the representations obtained from the pretrained model for learning to navigate using online RL?}
    \item \textit{How well can we plan using the BEV reconstructions from the pretrained model?}
    \item \textit{Does contrastive learning help learning good representations compared to an auxiliary task?}
    \item \textit{What are the performance benefits by adding ASC, TSC, and both?}
    \item \textit{How efficient and optimal is the navigation system when transferred to the Real-world setup?}
\end{enumerate}

\textbf{Policy Learning.} We performed RL experiments by deploying the frozen pretrained encoder and training a 1-layer policy in the CARLA simulator Fig. \ref{fig:6}. The task for the agent is to navigate to a goal destination using an RGB image ($o_t$, $\Delta g_t, \phi_t$). We accomplished this by training a policy employing the PPO algorithm \cite{schulman2017proximal}. The design of the reward function is rooted in proportionality to the number of waypoints the robot achieves to the designated goal point. In each timestep, the policy receives the current embedding of the observation $z_t$ concatenated with the directional vector pointing towards the waypoint tasked with producing a pair of (throttle, steer) values. We compared our method with VAE (reconstructing only the BEV image; Eqn. \ref{eqn:elbo}), TCN (trained using Eqn. \ref{eqn:tcn}), Random (Randomly initialized encoder and frozen), CLIP-RN50 \cite{DBLP:conf/icml/RadfordKHRGASAM21}. Note that, many of the prior works \cite{DBLP:conf/nips/AnandROBCH19, DBLP:conf/iclr/BurdaEPSDE19} have shown that randomly initialized and frozen encoders do learn decent features from an observation.

\textbf{Planning} We use TEB planner \cite{DBLP:journals/ras/RosmannHB17} to compute the action using an occupancy map (BEV reconstruction) to perform a task. Typically, occupancy map-based planners like TEB, use LiDAR data to compute the map of the environment and estimate a plan to perform the task, but in our case, we reconstruct the occupancy map using embedding obtained from RGB inputs. These maps are straightforward to compute in the case of our method and the VAE baseline, since these methods use a decoder. For the other baselines like the Random, CLIP and TCN encoder, we freeze the encoder and train the decoder to upsample the embeddings to estimate the BEV reconstruction. The results obtained for the planning task are shown in Fig. \ref{fig:6} as dotted lines.


\textbf{Quantitative Analysis} We evaluated the performance of our ResNet-50 model using the Validation dataset and the results are shown in Table \ref{fig:table_qual}. The performance of our perception model on both simulation and real-world dataset are compared to the baseline, which is a 6-way ResNet-50 classifier. Our perception model identifies the closest matching class for the output embedding. The baseline is a ResNet-50 model trained on a 6-class training dataset comprising 140,213 labelled FPV images. This proves that contrastive learning using BEV prediction enables better generalization, to out-of-domain data.

\textbf{Ablation Experiments for state checking} Following a similar approach, we used the Test dataset to evaluate the entire system. Apart from the accuracy also used Cross entropy (CE) and Mean Square error (MSE) to judge the quality of reconstructions by the LSTM model. These results are shown in Table \ref{fig:table_ablation}. Similar to the above experiments, we also used data from the unseen Town from the CARLA simulator to asses the predictions of our system. The metrics presented in this table exhibit a slight decrease compared to Table \ref{fig:table_qual}. This can be attributed to the increased presence of abnormal observations and higher ambiguity between classes within the time-series data obtained from the robot, as opposed to the manually collected and labelled dataset in the validation dataset.

\textbf{Evaluation on a Real-world system} We perform experiments on a Real-world robot, where the agent is tasked with navigating to a given destination location, using the pretrained \textit{ResNet} encoder and the trained policy in the Carla simulator. Success rates (SR) for planning experiments for our model are shown in Fig. \ref{fig:table_qual}. For both policy learning and planning, we specify the computation costs in Table. \ref{fig:table_flops}.



\section{Discussion and Future work}
In this paper we proposed a robust navigation system that is trained entirely in a simulator and frozen when deployed. We learn compact embeddings of an RGB image for Visual Navigation that are aligned with temporally closer representations and reconstruct corresponding BEV images. By decoupling the perception model from the control model, we get the added advantage of being able to pretrain the encoder using a set of observation sequences irrespective of the robot dynamics. Our system also consists of a memory module that enhances the robustness of the navigation system and is trained on an offline dataset from the simulator. Although our experiments in this paper are limited to data obtained through the simulator, one of the primary advantages of our methods is the ability to use additional simulator/real-world FPV-BEV datasets by aggregating with the current dataset.

\printbibliography
\end{document}